\documentclass[11pt]{llncs}
\usepackage[utf8]{inputenc}

\usepackage{authblk}
\usepackage{amssymb}
\usepackage{adjustbox}
\usepackage{float}
\usepackage{hyperref}
\usepackage{color,soul}
\usepackage{aliascnt}
\usepackage{amsmath}
\usepackage{textgreek}
\usepackage[a4paper, total={6.14in, 9.21in}]{geometry}
\usepackage{geometry}
\usepackage{ebgaramond,newtxmath,ebgaramond-maths}
\usepackage{glossaries}
\usepackage{chngcntr}
\counterwithin{figure}{section}
\usepackage{indentfirst}

\restylefloat{table}
\makeglossaries

\newacronym{ERP}{ERP}{Enterprise Resource Planning}
\newacronym{MES}{MES}{Manufacturing Execution Systems}
\newacronym{BFO}{BFO}{Basic Formal Ontology}
\newacronym{SUMO}{SUMO}{Suggested Upper Merged Ontology}
\newacronym{DOLCE}{DOLCE}{Descriptive Ontology for Linguistic and Cognitive Engineering}
\newacronym{GFO}{GFO}{General Formal Ontology}
\newacronym{P-PSO}{P-PSO}{Politecnico di Milano–Production Systems Ontology}
\newacronym{MRO}{MRO}{Manufacturing Reference Ontology}
\newacronym{MSO}{MSO}{Manufacturing Systems Ontology}
\newacronym{MSE}{MSE}{Manufacturing System Engineering}
\newacronym{GAN}{GAN}{Generative Adversarial Network}
\newacronym{GAAL}{GAAL}{Generative Adversarial Active Learning}

\title{Knowledge Modelling and Active Learning in Manufacturing}

\author{Jo\v{z}e M. Ro\v{z}anec\inst{1,2} \and Inna Novalija\inst{1} \and Patrik Zajec\inst{1}  \and Klemen Kenda\inst{1,2} \and Dunja Mladeni\'{c}\inst{1} }

\institute{
Jo\v{z}ef Stefan Institute, Jamova 39, 1000 Ljubljana, Slovenia\\
\email{\{joze.rozanec,inna.koval,patrik.zajec,klemen.kenda,dunja.mladenic\}@ijs.si}
\and
Jo\v{z}ef Stefan International Postgraduate School, Jamova 39, 1000 Ljubljana, Slovenia
}

\date{\today}

\begin{document}

\maketitle

\thispagestyle{plain}\pagestyle{plain}

\begin{abstract}
The increasing digitalization of the manufacturing domain requires adequate knowledge modeling to capture relevant information. Ontologies and Knowledge Graphs provide means to model and relate a wide range of concepts, problems, and configurations. Both can be used to generate new knowledge through deductive inference and identify missing knowledge. While digitalization increases the amount of data available, much data is not labeled and cannot be directly used to train supervised machine learning models. Active learning can be used to identify the most informative data instances for which to obtain users' feedback, reduce friction, and maximize knowledge acquisition. By combining semantic technologies and active learning, multiple use cases in manufacturing domain can be addressed taking advantage of the available knowledge and data.
\end{abstract}


\section{Introduction}\label{S:INTRODUCTION}
Digitalization enables collecting and storing data in a digital format. Digital data enables changes at the process, organization, and business domain levels \cite{parviainen2017tackling}. In manufacturing, it allows to achieve increased process efficiency (with lower performance variability and less unplanned downtime), a more efficient use of resources (e.g., lower energy consumption), increased safety and sustainability, product quality, and reduced product launch time\cite{sjodin2018smart,dalenogare2018expected,demartini2019digitalization}. Smart factories are built based on three principles\cite{sjodin2018smart}: cultivate digital people, introduce agile processes, and configure modular technologies. \cite{richter2017shop} identifies four digitalization challenges in manufacturing: how to digitally augment human work, enable worker-centric knowledge sharing, create self-learning manufacturing workplaces, and enable mobile learning. These challenges and benefits were recognized by several national and international initiatives (\href{https://trumpwhitehouse.archives.gov/wp-content/uploads/2018/10/Advanced-Manufacturing-Strategic-Plan-2018.pdf}{Advanced Manufacturing} (USA),  \href{https://www.plattform-i40.de/PI40/Navigation/EN/Home/home.html}{Industry 4.0} (Germany and the European Union)\cite{yang2021industry}, \href{http://english.www.gov.cn/2016special/madeinchina2025/}{Made in China 2025}, \href{https://www.meti.go.jp/english/press/2015/pdf/0123_01b.pdf}{New Robot Strategy} (Japan), \href{https://www.economie.gouv.fr/files/files/PDF/web-dp-indus-ang.pdf}{New Industrial France}, \href{https://innovateuk.blog.gov.uk/tag/high-value-manufacturing/}{High-Value Manufacturing} (UK), \href{https://www.industry.gov.au/data-and-publications/make-it-happen-the-australian-governments-modern-manufacturing-strategy/our-modern-manufacturing-strategy}{Make it Happen} (Australia)), and new paradigms created to realize them. Among such paradigms, we find Cyber-Physical Systems\cite{rajkumar2010cyber}, Digital Shadows, and Digital Twins\cite{grieves2014digital}. Cyber-Physical Systems were conceived as smart and embedded systems that result from the integration of physical and computational processes\cite{la2010service,negri2017review}. In Digital Shadows, the data flow is unidirectional (from the physical counterpart to the digital replica), while in Digital Twins, this flow is bidirectional (changes in the digital object can lead to changes in the physical object)\cite{kritzinger2018digital}. Multiple authors proposed enhancing the Digital Twins providing cognitive capabilities using a knowledge graph\cite{lu2020cognitive,rovzanec2021actionable}. Such technologies, along with the Internet of Things and Artificial Intelligence, bring added value into industrial value chains \cite{grangel:2019}.

To capture data in a digital form, sensors and software, such as as Enterprise Resource Planning (\acrshort{ERP}) or Manufacturing Execution Systems (\acrshort{MES}), are used. There are, however, many operational aspects and contextual information the employees are aware of that the sensors and the software systems mentioned above do not capture. Thus, it is essential to develop interfaces and mechanisms to gather such information while minimizing interaction friction with end-users. Some examples can be found in other domains, where to mitigate the knowledge gap, researchers developed conversational interfaces that identify missing knowledge and ask the users to provide it~\cite{preece2015sherlock,bradevsko2017curious}. In such a  context, semantic technologies and active learning can play a crucial role. Semantic technologies enable encoding domain knowledge (in ontologies and knowledge graphs) and provide means to perform inference (considering rules and logics)\cite{feng2017toward}. On the other side, active learning allows one to choose the most informative pieces of data to gather additional insights from experts.

This chapter discusses semantic knowledge representations (ontologies and knowledge graphs), the usage of active learning, and use cases in the industrial domain that benefit from both.

\section{Semantic knowledge representations}\label{S:SEMANTIC-KNOWLEDGE}
\subsection{Ontologies in manufacturing}\label{S:ONTOLOGIES}

Ontologies are explicit specifications of a conceptualization (an abstract, simplified view of the world) regarding objects, concepts, and entities, and the relationships between them \cite{gruber1993translation}. One of the main issues regarding knowledge management in the manufacturing domain is the wide range of concepts, problems, and configurations present\cite{garetti2015role}. A possible solution to this is the usage of semantic technologies. Ontologies provide a formal specification of a shared conceptualization in the domain of interest by defining concept hierarchies, taxonomies, and topologies\cite{ferrer2021comparing,gruber1995toward}. They provide information interoperability between different domains and enable reasoning. Among the ontology use cases in manufacturing mentioned in the literature, we find knowledge sharing and reuse in distributed manufacturing settings\cite{lin2011developing}, linking between product assemblies and manufacturing resources using manufacturing operations\cite{an2019ontology}, and production line processes\cite{zaletelj2018foundational}. Several ontologies were considered and developed in the manufacturing domain. Upper ontologies provide high-level concepts that can be extended to create domain-specific ontologies. Among the upper ontologies we find the Basic Formal Ontology (\acrshort{BFO})\cite{grenon2004snap}, Suggested Upper Merged Ontology (\acrshort{SUMO})\cite{niles2001towards}, Descriptive Ontology for Linguistic and Cognitive Engineering (\acrshort{DOLCE})\cite{gangemi2002sweetening}, General Formal Ontology (\acrshort{GFO})\cite{herre2010general}, Object-Centered High-level Reference Ontology (OCHRE) \cite{schneider2003designing}, Politecnico di Milano–Production Systems Ontology (\acrshort{P-PSO})\cite{garetti2012p}, Manufacturing Reference Ontology (\acrshort{MRO})\cite{usman2013towards}, Manufacturing Systems Ontology (\acrshort{MSO})\cite{negri2015ontology}, and the Manufacturing System Engineering (MSE) ontology model\cite{lin2007manufacturing}. 

The \acrshort{BFO} ontology attempts to model time and space. To that end, it divides entities into two disjoint categories: continuants (something that exists at a point in time) and occurrents (something that is realized in time, e.g., processes and events). \acrshort{SUMO} is considered the largest formal public ontology, mapping the whole WordNet lexicon. It divides entities into two disjoint categories: physical (represents objects and processes) and abstract (represents sets, propositions, quantities, and attributes). The upper ontology is complemented with the MId-Level Ontology (MILO), and domain-specific ontologies are developed on top of them. \acrshort{DOLCE} was developed to capture ontological categories underlying natural language and human commonsense. Their focus is to describe categories as cognitive artifacts as represented in human perception, rather than in the intrinsic nature of the world. The ontology divides entities into two categories: \textit{endurants} (continuants) and \textit{perdurants} (occurrents). A mapping between \acrshort{BFO} and \acrshort{DOLCE} was proposed in \cite{temal2010mapping}. \acrshort{GFO} provides a different worldview, dividing entities into two categories: presential and process. \textit{Presentials} refer to entities that are entirely present at a given point in time. To model how \textit{presentials} acquire different values in time but remain the same entity, they refer to \textit{persistents} (a specific universal representing the \textit{presentials}). The processes represent functions that have a temporal extension and cannot be wholly present only at a given point in time. The persistent-presential aspects are discussed in OCHRE under the terms of \textit{thick} and \textit{thin objects}, where the thick objects refer to aspects that change over time, while the thin object refers to core aspects that remain the same through time. While the ontology distinguishes between \textit{endurants} and \textit{perdurants}, it does so by modeling participation as a special case of parthood and avoids assuming two separate domains. \acrshort{P-PSO} was designed as a meta-model to describe the manufacturing domain from an object-oriented perspective. When doing so, it considers three aspects of the manufacturing setting: physical (entities' material definition), technological (system functional view), and control (production operation procedures) aspect. The \acrshort{MSO} evolves the \acrshort{P-PSO}, addressing a wider domain, built with a different purpose, and providing a different approach to the control and visualization aspects. Regarding the domain, high-level classes are defined to address all types of industry, and specific classes are defined as specializations of such high-level objects. In particular, the ontology extends the scope to logistics and the process industry.
In contrast to the \acrshort{P-PSO}, which provides a general taxonomy but does not define a specific usage, \acrshort{MSO} was designed for production system control. Finally, regarding control and visualization aspects, \acrshort{P-PSO} defines entities and relationships to be considered for manufacturing control. In contrast, \acrshort{MSO} provides definitions at a conceptual level, assuming the ontology only interacts with different software placing its interest on the outcomes, without the need to represent the inner design and working of the software service. The \acrshort{MRO} was designed as an upper manufacturing ontology, defining the terminology based on existing standards. A different approach was adopted by the \acrshort{MSE} ontology model, which provides a model to support information autonomy and facilitate information exchange between inter-disciplinary engineering design teams while leaving to each team freedom to adopt their terminology. The aforementioned upper ontologies are considered when creating domain-specific ontologies.

Manufacturing always relates to a specific product, and authors developed specific ontologies to describe them. \cite{vegetti2005product} developed PRONTO (PRoduct ONTOlogy), which defines concepts, relationships, and axioms mainly related to the manufactured products' structure. The ontology considers raw materials, how those are assembled into a product, and derivative products. \cite{panetto2012onto} noticed that ontologies and standards aim to facilitate a common grounding by sharing expert knowledge and finding agreement on a particular domain. They developed the ONTO-PDM ontology based on existing models\footnote{IEC 62264 models are: Product Definition, Material, Equipment, Personnel, Process Segment, Production Schedule, Production Capability, and Production Performance} from the IEC 62264 standard.

Products cannot be developed without a specific manufacturing process. \cite{schlenoff2000process} developed the Process Specification Language (PSL) ontology to describe manufacturing processes throughout the manufacturing life cycle. \cite{zaletelj2018foundational} developed and applied a meta-model to describe a material-processing production line, which supports defining the behavior of an entity over time through state transitions. \cite{jarvenpaa2019development} developed the Manufacturing Resource Capability Ontology (MaRCO), to describe the capabilities of manufacturing resources, concentrating solely on machines and tools, so that can be used to support semi-automatic system design and auto-configuration of production systems. Another effort to describe products, production processes and resources is the P2 ontology, developed by \cite{jasko2020development}. \cite{borgo2007foundations} describe a manufacturing ontology-based on the \acrshort{DOLCE} ontology and the Adaptive Holonic Control Architecture for distributed manufacturing systems (ADACOR)\cite{leitao2005adacor}, that describes manufacturing scheduling and control operations. Another view on scheduling was developed by  \cite{vsormaz2019simpm}, who introduced the SIMPM (Semantically Integrated Manufacturing Planning Model) ontology, modeling manufacturing planning task according to time, variety, and aggregation. \cite{lu2013ontology} presents the Supply Chain Operations (SCOR) ontology to facilitate the interoperation between applications involved in the supply chain. A product data model in a cloud manufacturing context was developed by \cite{lu2019manuservice}. Finally, additive manufacturing was subject of several ontologies\cite{sanfilippo2019ontology,ali2019product}.

Another relevant aspect to the manufacturing domain is the sensors, which enable data gathering. OntoSensor\cite{russomanno2005building} aims to provide a broad knowledge base of sensors for query and inference, based on the SensorML standard\footnote{SensorML is an approved Open Geospatial Consortium standard. More details are available at \url{https://www.ogc.org/standards/sensorml}}. In the same line, \cite{neuhaus2009semantic} proposed an ontology to describe sensors' capabilities and operations.  \cite{hu2007ontology} developed WISNO (Wireless Sensor Networks Ontology) to deduce high-level information from low-level, implicit context and checking ontologies' consistency. \cite{compton2009reasoning} describes an ontology to characterize sensor capabilities and properties as the composition of their building blocks through three description levels: domain concepts, abstract sensor properties, and concrete properties.

\subsection{Methodologies for ontology design in manufacturing}\label{S:ONTO-METHODOLOGIES}
Rarely an ontology satisfies all the requirements and frequently needs to be extended, or new ontologies need to be developed to cover a new domain. Multiple methodologies were described in the literature to build an ontology. While each methodology has a different emphasis, there is consensus on most steps to be followed:
\begin{itemize}
    \item identify the problem to be solved, opportunity areas, and a potential solution\cite{uschold1995towards,uschold1996building,fernandez1997methontology}
    \item decide on the formality level required\cite{uschold1996building}
    \item define the problem, scope and competency questions\cite{kim1995ontology}
    \item elicit required knowledge from multiple sources. Identify key concepts and relationships. Identify terms that refer to the concepts and relationships. \cite{uschold1995towards,uschold1996building,fernandez1997methontology}. When doing so, consider the MIREOT guidelines\cite{courtot2011mireot}.
    \item evaluate against a frame of reference\cite{uschold1995towards,fernandez1997methontology}
\end{itemize}

Specific methodologies were developed to guide the ontologies' construction in the manufacturing domain. \cite{ahmed2007methodology} proposed a six-stage methodology: identify root concepts of taxonomies, identify existing taxonomies, create taxonomies, application test, build terms thesaurus, and refine the integrated taxonomy.  \cite{li2009methodology} also defined a six-step methodology but provided a different procedure: specification (determine scope and granularity), conceptualization (acquire knowledge), formalization (structure acquired knowledge), population (convert acquired knowledge into frame-based representation), evaluation (validate accuracy and completeness), and maintenance (update the ontology once established). A different approach was developed by \cite{lin2011developing}, who combined the Unified Modelling Language and the Object Constrained Language to translate entities from a software object model to ontology entities. \cite{ameri2012systematic} developed a four-step methodology using a Simple Knowledge Organization System (SKOS) framework to develop a thesaurus of concepts, to then identify relevant classes and provide logical constraints and rules. \cite{kiritsis2012design} suggests a three-step methodology, inspired in \cite{suarez2010gontt}, that requires an ontology requirements specification (purpose, scope, and ontology requirements analysis), an analysis of existing resources (reuse ontological and non-ontological resources), and a conceptualization and formalization. A similar approach was developed by \cite{chang2010development}, with an emphasis on manufacturing design. \cite{akmal2013methodology} defined a nine-step methodology for developing process ontologies. First, it requires defining the project's purpose and scope, identifying potential classes and formal attributes, and writing them down to a context table. Concepts and subconcepts should be drafted to a lattice, which is used to resolve inconsistencies, and then converted to a class hierarchy. The last steps correspond to integrate the hierarchy with some upper ontology and the classes formally defined through axioms and relationships. Finally, \cite{hildebrandt2020ontology} envisions a different scenario, creating an ontology building methodology for Cyber-Physical Systems in the manufacturing domain. The methodology consists of three steps: ontology requirements specification (based on project requirements), lightweight ontology building (considering requirements, information resources, and other lightweight ontologies), and heavy-weight ontology building (taking into account the lightweight ontology and ontology design patterns).

\subsection{Knowledge graphs in manufacturing}\label{S:KNOWLEDGE-GRAPH}

Among the rich literature describing knowledge graphs, there is no agreed unique definition for them. The knowledge graphs are built upon the idea that graphs can be used to capture knowledge. Nodes are used to define abstractions and instantiate entities, which can be linked with edges, representing relationships \cite{ji2021survey}.  They can be either domain-specific or domain independent \cite{paulheim2017knowledge}. Many implementations constrain the edges in knowledge graphs according to some schema or ontology\cite{noy2019industry}, providing a formal concepts' definition. \cite{hogan2020knowledge} provides a comprehensive introduction to knowledge graphs, discussing data models, schemas, deductive and inductive techniques, quality dimensions, refinement methods, and prominent open and enterprise knowledge graphs. Deductive inference can be used to derive new knowledge from existing data and rules known \textit{a priori}. Inductive knowledge, on the other side, is acquired by generalizing patterns from input observations, either using supervised or unsupervised methods.  Knowledge graph refinement attempts to identify wrong information in the graph (ensure that it is free of error) and complete missing information (satisfy completeness)\cite{paulheim2017knowledge}. Such tasks can benefit from knowledge graph embeddings, which reduce nodes and edges to continuous vector spaces while preserving the inherent graph structure\cite{wang2017knowledge}. When assessing the quality of a knowledge graph, \cite{hogan2020knowledge} highlights four quality dimensions: accuracy (the extent to which the knowledge graph represents the real-world domain), coverage (avoid the omission of elements that are relevant to the specific domain), coherency (conformity to formal schema or ontology), and succinctness (avoid irrelevant data). \cite{farber2018linked} describes a wide range of quality metrics, classifying them in four quality categories described by \cite{wang1996beyond}: intrinsic data quality (quality of data on its right, regardless of the use case), contextual data quality (assessed concerning the task at hand), representational data quality (relates to the format and meaning of the data), and accessibility data quality (relates to how data can be accessed, considering accessibility, licenses, and interlinking).

Knowledge graph implementations can adopt one of three assumptions: open world, locally closed world\cite{dong2014knowledge}, or closed world assumption. Open world assumption considers that a statement can be true irrespectively of whether it is known to be true since there is much unknown information compared to the encoded knowledge. The local-closed world assumption considers the knowledge representation is locally complete. The truth regarding a statement can be determined as long as the set of existing object values for a subject and predicate are not empty. Finally, the closed-world assumption assumes that only statements known to be true can be true.

Manufacturing knowledge is gaining an increasing amount of attention \cite{he2019manufacturing}. The use of knowledge graphs to model it was reported in multiple scenarios. \cite{yan2020knowime} describe building and using a knowledge graph to integrate information of products and equipment obtained from heterogeneous data sources. The knowledge graph is a cornerstone to an intelligent manufacturing equipment information system. \cite{lv2020supplier} report encoding purchase records data in a supply chain knowledge graph and use embeddings to recommend the best suppliers for the purchase demand. \cite{ding2019robotic} use natural language processing to extract disassembly data (entities and the nature of components) and then encode it in a knowledge graph, which helps to acquire, analyze and manage disassembly knowledge. A different purpose is envisioned by \cite{munir2020knowledge}, who integrate semantic information of the workers with temporal profiling information, and facial recognition. Finally, \cite{rivas2020unveiling,grangel2019knowledge} describe a knowledge graph to integrate information regarding Industry 4.0 standards and standardization frameworks. Using graph embeddings, they can detect standards relatedness, identify similar standards and unknown relations.

\section{Active learning}\label{S:ACTIVE-LEARNING}
Active learning is a field of machine learning that studies how to select unlabeled data samples and query an information source to label the selected samples\cite{kumar2020active,schroder2020survey,budd2021survey}. The underlying assumption is that unlabeled data is abundant, and labeling resources are scarce. Therefore, it is necessary to devise mechanisms that enable the identification and selection of samples with a higher information potential. The promise to reduce the amount of data required to train new models has driven increased interest in active learning in the academic community. At the same time, the adoption remains low in industry\cite{samsonov2019more}.

Three different active learning scenarios are described in the literature\cite{settles2009active}: membership query synthesis, stream-based selective sampling, and pool-based active learning. In membership query synthesis\cite{angluin1988queries,schumann2019active} an algorithm creates its instances (queries) from an underlying distribution to ask the expert if the instance corresponds to a particular label. Stream-based selective sampling considers one unlabeled instance at a time, evaluating its informativeness against the query parameters. The learner decides whether to query the teacher or assign the label by itself. Finally, in pool-based sampling, unlabeled instances are drawn from the entire data pool and assigned an informative score. Most informative instances are selected, and their labels requested. While most methods rely on model uncertainty and clustering to choose the unlabeled examples\cite{cohn1996active,fu2013survey}, new approaches were developed based on adversarial sampling, Bayesian methods, and weak supervision. 
\cite{zhu2017generative} introduced a new approach to active learning (Generative Adversarial Active Learning (\acrshort{GAAL})) by leveraging Generative Adversarial Networks (\acrshort{GAN}). The purpose of the \acrshort{GAN}s is to generate informative instances based on a random sample of unlabeled instances close to the decision boundary. \cite{mahapatra2018efficient} evolved this concept generating synthetic data with a conditional \acrshort{GAN}, which learns to create a specific instance leveraging additional data regarding the desired target label. \cite{sinha2019variational} introduced the variational adversarial active learning, sampling instances using an adversarially trained discriminator to predict whether the instance is labeled or not based on the latent space of the variational auto-encoder. Since the sampling ignores the instance labels, the discriminator can end up selecting instances that correspond to the same class, regardless of the proportion of labeled samples of such class. To solve such an issue, \cite{ebrahimi2020minimax} developed a semi-supervised minimax entropy-based active learning algorithm that leverages uncertainty and diversity in an adversarial manner. Another approach was developed by \cite{mayer2020adversarial}, who, instead of uncertainty sampling, used a \acrshort{GAN} to generate high entropy samples and retrieve similar unlabeled samples from available data to acquire the corresponding labels.
A variation to \acrshort{GAAL}, and based on previous work by \cite{houlsby2011bayesian}, \cite{tran2019bayesian} developed a Bayesian generative active deep learning approach, performing a joint training of the generator (a variational autoencoder) and the learner, which requires smaller sample sizes and a single training stage. Different approaches were developed by \cite{boecking2020interactive,awasthi2020learning,biegel2021active}, who explored using active learning in a weak supervision setting.

Despite the wide range of active learning approaches, there is currently a research void regarding the use of active learning in the manufacturing domain\cite{meng2020machine}. It was successfully applied to predict the local displacement between two layers on a chip in the semi-conductor industry\cite{van2018active}, for automatic optical inspection of printed circuit boards\cite{dai2018towards}, to improve the predictive modeling for shape control of composite fuselage\cite{yue2020active}, and in multi-objective optimization\cite{lv2019surrogate}.

\section{Use cases and open challenges}\label{S:USE-CASES}

Semantic technologies and active learning can be used to identify missing knowledge. Within the semantic domain, \cite{pradel2020question} proposed a typology of missing knowledge, identifying three types of missing knowledge: \textit{abstraction dimension} (how the knowledge is contained inside the KG structure), \textit{terminological knowledge} (how to map terms to concepts), and \textit{question-answering dimension} (how the lack of knowledge affects the answering process). \cite{preece1993new} proposed to frame the missing knowledge problem as an anomaly detection problem, where they use a heuristic to identify missing knowledge in system rule bases. They take into account user input during the inference process for items that are considered \textit{askable}. \cite{xu2018provenance} suggested an approach based on first-order logic and dual polynomials. They use triples consisting of a question, answer, and a label that can indicate if the answer is missing or wrong. For missing answers, they developed heuristics to create potential answers that comply with a closed world setting. \cite{franklin2011crowddb} proposed developing an interface to issue SQL queries that can target either a relational database or crowdsource certain operations, such as find new data or perform non-trivial comparisons. The authors consider that while many operations can be successfully completed with data within a database, humans can assist with operations such as gathering missing data from external sources, moving towards an open-world assumption. 

\cite{waltinger2013usi} combined ontologies with natural language processing to develop a question answering interface that enabled users to access available underlying data sources. Among other results, the authors highlighted how such a system provided a positive experience to the users, doubling user retention. \cite{wang2017knowledge} describes the use case, using a knowledge graph to simplify question answering by organizing them in a structured format. \cite{bordes2014open} tackles question answering by creating vector embeddings of questions and knowledge graph triples so that the question vectors end up close to the answer vectors. A different approach was considered by \cite{bradevsko2017curious}, who developed \textit{Curious cat}. This application leverages a semantic knowledge base and user's contextual data for knowledge acquisition through question-answering. A similar knowledge acquisition approach for the manufacturing domain was envisioned by \cite{rovzanec2021xaikg}, who developed an ontology to model user feedback based on a given forecast and provided explanations. Following the need to augment human work with digital technologies and provide personalized information at the shop-floor level\cite{kagermann2015change}, \cite{zajec2021towards} developed a smart assistant for manufacturing. The smart assistant creates directive explanations for the users by using heuristics and domain knowledge. The application tracks user's implicit and explicit feedback regarding local forecast explanations, enabling application-grounded evaluations. Though the authors tested their approach on the demand forecasting use case, the application can be extended to other use cases. Other relevant use cases are the usage of semantic technologies to build a decision support system\cite{alkahtani2019decision}, automatically identify opportunities to enhance production scenarios\cite{giovannini2012ontology}, and intelligent condition monitoring of manufacturing tasks\cite{cao2019smart}.

Though little research reports on the usage of active learning in manufacturing\cite{meng2020machine}, we consider it can be widely applied in this domain. By selecting the most valuable instances to the system, it helps to minimize friction towards the end-user and collect valuable data~\cite{elahi2016survey}. Active learning can also increase the diversity of recommendations\cite{yang2020unifying}. This approach can be used in applications recommending decision-making options to balance usual recommendations and decision-making options requiring more user feedback (more labeled instances) to enhance the underlying recommender system. Other relevant active learning use cases to manufacturing can be anomaly and outlier detection\cite{pelleg2004active,abe2006outlier,liu2019generative}.

In the European Horizon 2020 project STAR (Safe and Trusted Human Centric Artificial Intelligence in Future Manufacturing Lines), knowledge modeling and active learning are used to gather locally observed collective knowledge regarding operations in the manufacturing lines and provide accurate context, relevant data, and decision-making options to the users. Among relevant use cases to the project are production planning (to gather additional context regarding downtimes and anomalies in production), optical quality inspection (to learn from images of defective parts), and logistics (to learn logisticians' decision-making based on available options).

\section{Conclusion}\label{S:CONCLUSION}

Semantic technologies provide means to encode domain knowledge and enable deductive inference through reasoning engines. In this work we presented many upper-level and domain-specific ontologies from the manufacturing domain, and upon which new ontologies can be built. We also described multiple methodologies used to guide the ontology creation process, some of them specific to the manufacturing domain.

Semantic technologies can be leveraged for knowledge acquisition. Missing knowledge detection can be linked to a question-answering interface to gather required knowledge from the users. Similarly, active learning can be used can identify the most informative data instances and ask the users for feedback. This enables to gradually increase the dataset and its information density, which can be leveraged to train machine learning models, and enhance their performance. While little scientific literature reports on the usage of active learning in the manufacturing domain, multiple use cases can benefit from it, such as anomaly detection in production planning, optical quality inspection, and the recommendation of decision-making options.

\section*{Acknowledgements}
This work has been carried out in the H2020 STAR project, which has received funding from the European Union's Horizon 2020 research and innovation programme under grant agreement No. 956573.

\bibliographystyle{ieeetr}
\bibliography{main}

\printglossary[title=Abbreviations]

\end{document}